\documentclass{article} 
\usepackage{iclr2020_conference,times}

\usepackage{amsmath,amsfonts,bm}

\def\eqref#1{equation~\ref{#1}}

\def\1{\bm{1}}

\DeclareMathAlphabet{\mathsfit}{\encodingdefault}{\sfdefault}{m}{sl}
\SetMathAlphabet{\mathsfit}{bold}{\encodingdefault}{\sfdefault}{bx}{n}

\usepackage{graphicx}
\usepackage{wrapfig}
\usepackage[hidelinks]{hyperref}
\usepackage{url}
\usepackage{color}
\usepackage{endnotes}
\let\footnote=\endnote

\title{Field-Level Crop Type Classification with $k$ Nearest Neighbors: A Baseline for a New Kenya Smallholder Dataset}

\author{Hannah R. Kerner, Catherine Nakalembe, \& Inbal Becker-Reshef  \\
Department of Geographical Sciences\\
University of Maryland at College Park\\
College Park, MD 20742, USA \\
\texttt{\{hkerner,ireshef,cnakalembe\}@umd.edu}
}

\iclrfinalcopy 
\begin{document}

\maketitle

\begin{abstract}
Accurate crop type maps provide critical information for ensuring food security, yet there has been limited research on crop type classification for smallholder agriculture, particularly in sub-Saharan Africa where risk of food insecurity is highest. Publicly-available ground-truth data such as the newly-released training dataset of crop types in Kenya (\href{https://www.mlhub.earth/}{Radiant MLHub}) are catalyzing this research, but it is important to understand the context of when, where, and how these datasets were obtained when evaluating classification performance and using them as a benchmark across methods. In this paper, we provide context for the new western Kenya dataset which was collected during an atypical 2019 main growing season and demonstrate classification accuracy up to 64\% for maize and 70\% for cassava using $k$ Nearest Neighbors---a fast, interpretable, and scalable method that can serve as a baseline for future work.
\end{abstract}

\section{Introduction}
\label{intro}

Accurate crop type maps are essential for estimating and forecasting crop production, conditions, and yields for major food-producing countries as well as those at risk for food insecurity \citep{Becker-reshef2020}. While crop type classification methods for large-scale farming regions such as the United States are significantly advanced, methods for smallholder farming regions such as Eastern Africa have had less attention and thus remain nascent \citep{Fritz2010,Lobell2018}. One reason for this is a lack of publicly-available ground-truth datasets for smallholder farms that can be used to train machine learning models. The Radiant Earth Foundation recently released new datasets to catalyze development of crop type classification methods for smallholder regions. One dataset contains ground-truth field boundary and crop type labels (co-registered to Sentinel-2 satellite images) that were collected during a field campaign by Plant Village in western Kenya in 2019, which was atypical due to severe weather conditions. If this and other ground-truth datasets are to be used as benchmarks for smallholder crop type classification methods in general, it is important to understand how representative the samples are of a typical growing season or smallholder crop type classification problem. 

 Most machine learning methods for crop type classification in smallholder-dominated regions involve random forests \citep{Jin2019,Lambert2018,Lebourgeois2017,Aguilar2018} and support vector machines (SVMs) \citep{Aguilar2018}. While recent work has proposed deep learning methods \citep{Rustowicz2019}, application of such approaches to crop type classification in smallholder regions has been limited by a lack of available labeled training data in these regions. As new public datasets spur development of models with varying complexity, interpretability, and computational requirements, it is important to understand the baseline performance that should be expected given the limitations of the dataset. In this paper, we provide an interpretation of the new Kenya crop type dataset in the context of the atypical 2019 growing season and demonstrate baseline classification performance for this dataset using $k$ Nearest Neighbors, chosen for its speed, interpretability, and scalability. 

\section{Method}
\label{method}

\subsection{Data and study area}
\label{dataset}
\begin{figure}[t]
\centering
\includegraphics[width=1.0\linewidth]{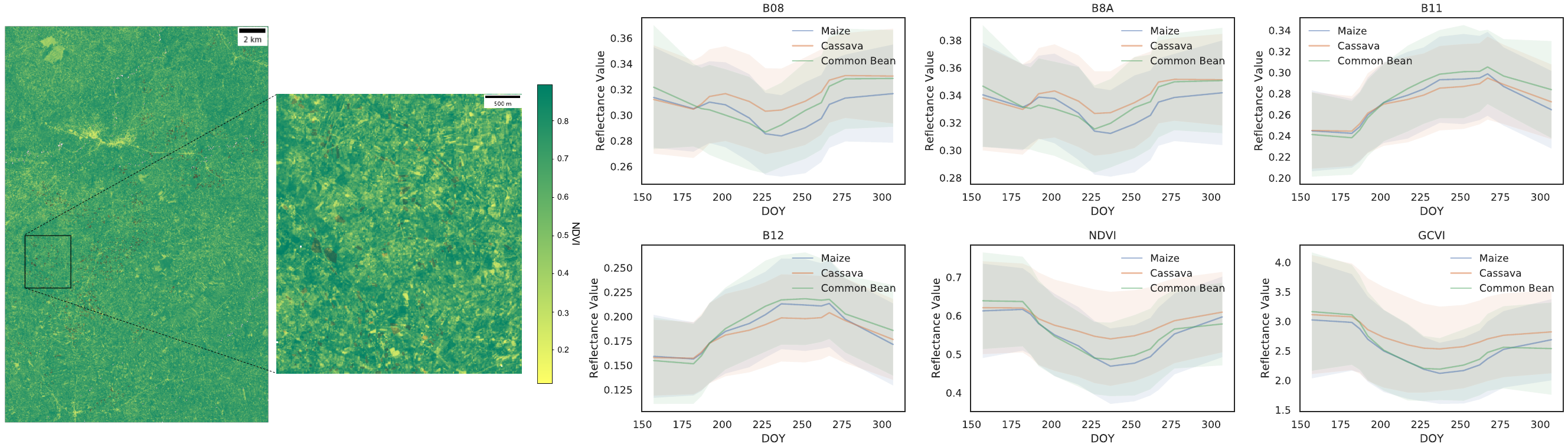}
\caption{Left: NDVI in region 03 of Sentinel-2 data with inset and field boundaries overlaid in transparent red (image acquired July 1, 2019). Right: NDVI time series values for select Sentinel-2 bands (842 nm NIR, 865 nm narrow-band NIR, 1610 nm SWIR, and 2190 nm SWIR) and vegetation indices (NDVI and GCVI).}
\label{fig:bandsep}
\end{figure}

For this study, we used the crop type dataset published as part of the Radiant Earth Foundation Crop Classification using Earth Observations Challenge.\footnote{\url{https://www.cv4gc.org/cv4a2020}} This dataset contains field boundaries and crop types for $>$3,000 fields in western Kenya collected during the 2019 main growing season. There are a total of seven crop type classes: three pure classes---maize, cassava, and common bean---and four intercropped classes---maize/common bean, maize/cassava, maize/soybean, and cassava/common bean. The dataset also provides Sentinel-2 multispectral observations over four regions for 13 days (spaced unevenly) between June 6-November 3, 2019 (52 observations total), with 10 m/pixel spatial resolution. Figure \ref{fig:bandsep} (left) shows NDVI for an image acquired on July 1, 2019 overlaid with field boundary labels for one of the four regions. To pre-process the data, we 1) masked out pixels with cloud probability $> 0$; 2) computed median time series within the field boundary (per-band); 3) masked out timesteps outside of 5th and 95th percentiles (per-band); and 4) filled gaps using Savitzky-Golay smoothing \citep{Chen2004}. Step 3 is necessary because the Sentinel-2 cloud probability layer does not reliably identify all clouds in the scene. To evaluate which band(s) would enable greatest separability between crop classes, we plotted the mean and standard deviation time series within the pure-crop classes (maize, cassava, common bean) for select bands and popular vegetation indices (NDVI and GCVI) in Figure \ref{fig:bandsep} (right). This shows that the crop classes have high intra-class variance in all bands, but that the NDVI and GCVI time series have the greatest inter-class variance. Since NDVI and GCVI provide similar separability, we used NDVI for our model input since it is more widely used. 

\begin{figure}[t]
\centering
\includegraphics[width=\linewidth]{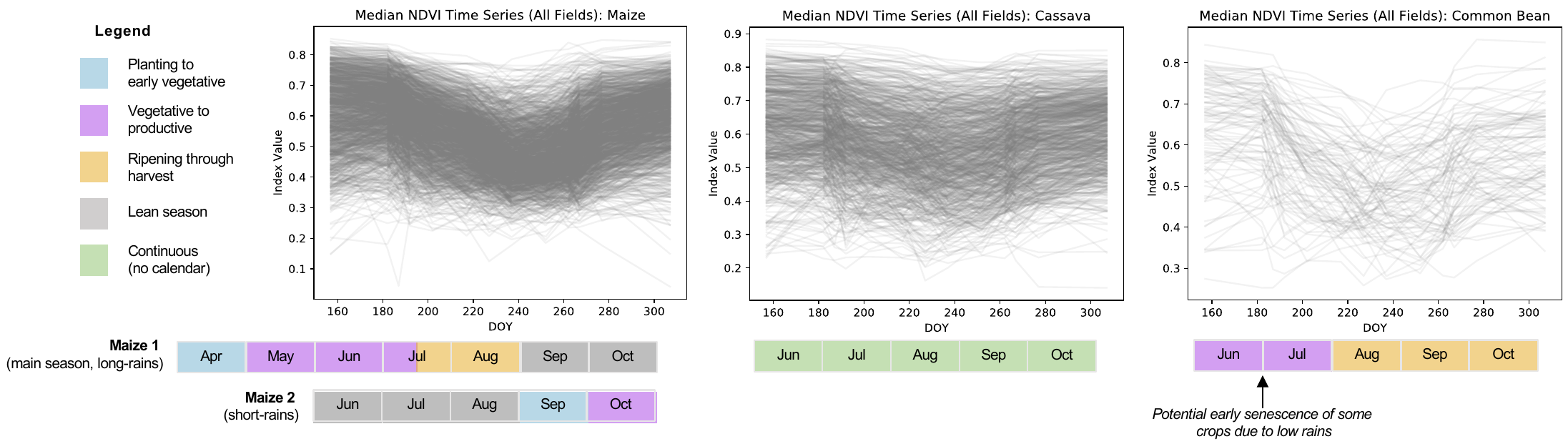}
\caption{Median NDVI time series within season for all fields in the dataset for single-crop types. Timelines show expected crop calendars for each crop type annotated with the stage in the crop calendar.}
\label{fig:ndvi_timeseries}
\end{figure}

Of the fields in the training dataset, 1,462 are maize, 829 are cassava, 487 are maize/common bean, 172 are maize/cassava, 160 are maize/soybean, 98 are common bean, and 78 are cassava/common bean. Figure \ref{fig:ndvi_timeseries} shows the median NDVI time series for pure fields by crop type. Below the time series, we show crop calendars we constructed based on the Crop Monitor for Early Warning reports (published monthly) and Kenya National Crop Conditions Bulletins (published approximately bi-monthly). Maize 1 is the ``long-rains'' primary growing season in western Kenya. Planting for long-rains maize typically begins in March but was delayed in 2019 due to cyclones and below-average rains (as much as 100\% below average in some areas); nevertheless, planting was completed by the end of April.\footnote{Crop Monitor for Early Warning, May 2019: \url{https://cropmonitor.org/index.php/2019/05/09/crop-monitor-for-early-warning-may-2019/}}\footnote{Kenya Crop Conditions Bulletin, March 2019: \url{http://www.kilimo.go.ke/wp-content/uploads/2019/05/National-Crop-Conditions-Bulletin_APRIL-2019.pdf}} This is consistent with the maize time series in Figure \ref{fig:ndvi_timeseries}, which shows a peak around June/July, indicating the culmination of the vegetative to productive stage. The second peak in the maize NDVI time series is likely due to regrowth (e.g., of grasses or weeds) and growth of Maize 2, which is the ``short-rains'' secondary maize crop. Maize 2 was reportedly in the planting to early vegetative state as of September 28\footnote{Crop Monitor for Early Warning, October 2019: \url{https://cropmonitor.org/index.php/2019/10/02/crop-monitor-for-early-warning-october-2019/}} and nearing the end of the vegetative to productive stage by the end of November.\footnote{Kenya Crop Conditions Bulletin, November 2019: \url{http://www.kilimo.go.ke/wp-content/uploads/2020/01/Kenya-Crop-Conditions-Bulletin-November-2019.pdf}} Cassava is a perennial woody shrub that requires at least 6-12 months to produce a first crop, and its leaves reach maximum area 4-5 months after planting. Cassava is tolerant to long dry periods and can remain in the ground for over two years. These factors cause lower dynamic range in NDVI throughout the season compared to other crops, which is evident in Figure \ref{fig:ndvi_timeseries}. Common beans are typically planted at the same time as maize, but late rains cause delayed establishment of beans by one month.\footnote{Kenya Crop Conditions Bulletin, May 2019: \url{http://www.kilimo.go.ke/wp-content/uploads/2019/06/National-Crop-Conditions-Bulletin_May-2019-FinalEdition2.pdf}} By end of July, harvesting was ongoing in most areas but some areas experienced crop failure due to wet conditions after July rains\footnote{\label{kccb_july}Kenya Crop Conditions Bulletin, July 2019: \url{http://www.kilimo.go.ke/wp-content/uploads/2019/08/National-Crop-Conditions-Bulletin_July-2019-Edited.pdf}}. We also speculate that the sharp drop in NDVI observed in Figure \ref{fig:ndvi_timeseries} between June and July could be due to early senescence of bean crops due to failed rainfall at the start of the season.\textsuperscript{\ref{kccb_july}}

\subsection{K-Nearest Neighbors Crop Type Classification}
$k$ Nearest Neighbors ($k$NN) is a supervised classification method that assigns unlabeled examples with the label of the majority class of the $k$ examples closest to the test example in the feature space \citep{Mucherino2009}. We used this method to evaluate baseline classification performance for the Kenya dataset because it is easily interpretable, fast, and scalable to large areas. It is also computationally simple, which is important because high computational requirements can be a barrier to stakeholder uptake. While other methods such as deep neural networks and random forests have been popular in prior work (e.g., \citet{Rustowicz2019,Jin2019}), these methods are more complex, computationally intensive, less interpretable, and (particularly in the case of neural networks) susceptible to overfitting with small datasets. We used the cosine similarity metric to compute pairwise distances between examples. We did not see improvement in classification accuracy by reducing the dimensionality of the NDVI time series (e.g., using principal component analysis), thus we used the 13-date field-level NDVI time series as the input representation. This representation has the benefit of being immediately interpretable, since classification is based on a straightforward measure of similarity between time series. 

\section{Results and Discussion}
\label{experiments}

\begin{table}[t]
\caption{$k$NN classifier accuracy for experiments including maize (Ma), cassava (Ca), maize/common bean (Ma/CB), maize/cassava (Ma/Ca), maize/soybean (Ma/So), common bean (CB), cassava/common bean (Ca/CB), and overall accuracy (OA). \textbf{Bold}: highest per-class accuracy. }
\label{results}
\begin{center}
\begin{tabular}{lllllllll}
\multicolumn{1}{c}{\bf Accuracy}  &\multicolumn{1}{c}{\bf OA} &\multicolumn{1}{c}{\bf Ma} &\multicolumn{1}{c}{\bf Ca}  &\multicolumn{1}{c}{\bf Ma/CB}  &\multicolumn{1}{c}{\bf Ma/Ca} &\multicolumn{1}{c}{\bf Ma/So} &\multicolumn{1}{c}{\bf CB} &\multicolumn{1}{c}{\bf Ca/CB} 
\\ \hline \\

2 crops ($m=829, k=9$) & 68.4 & 64.3 & \textbf{70.3} \\
3 crops ($m=487, k=15$) & 49.0 & 44.0 & \textbf{58.1} & 45.0 \\
4 crops ($m=172, k=9$) & 37.2 & 44.2 & \textbf{51.1} & 36.6 & 16.8 \\
5 crops ($m=160, k=11$) & 30.3 & 36.9 & \textbf{54.4} & 25.6 & 21.3 & 13.1 \\
6 crops ($m=98, k=15$) & 23.6 & 27.7 & \textbf{47.8} & 18.3 & 24.4 & 21.3 & 11.2 \\
7 crops ($m=78, k=13$) & 18.0 & 24.3 & \textbf{42.3} & 18.3 & 14.1 & 10.3 & 6.3 & 10.1 \\

\end{tabular}
\end{center}
\end{table}

One limitation of $k$NN is that it can be sensitive to class imbalances since the $k$ nearest neighbors of a test point would be more likely to belong to the majority class. We performed six experiments designed to assess $k$NN performance with balanced classes while using as much of the available 
\begin{wrapfigure}{r}{0.5\textwidth}
  \begin{center}
\includegraphics[width=1.0\linewidth]{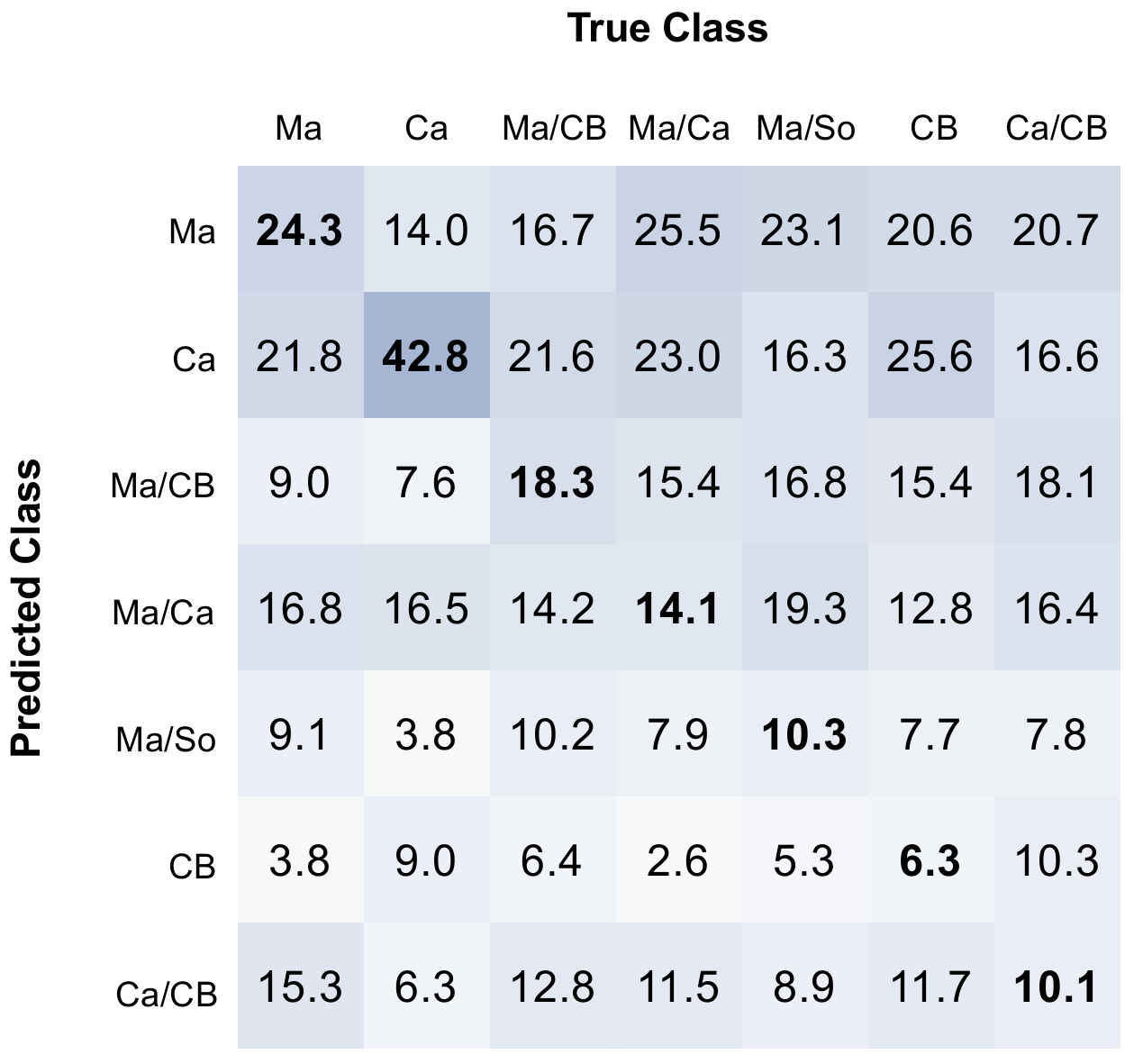}
  \end{center}
\caption{Confusion matrix for 7-crop $k$NN.}
\label{fig:confusion_matrix}
\end{wrapfigure}
training examples as possible. In the first experiment, we included only the two classes with the most examples (maize and cassava) and undersampled the maize examples to have the same frequency as cassava. In each subsequent experiment, we added the next-most frequent class to the dataset and undersampled the majority classes to be the same size as the minority class. To undersample, we randomly selected $m$ examples from the majority class(es) where $m$ is the minority class size (using the same random seed for all experiments). Thus in every experiment, all classes have the same number of examples. In Table \ref{results}, we report the average overall and per-class classification accuracy from 5-fold cross validation for each experiments. We used the value of $k$ that gave highest accuracy for odd $k \in [1, 19]$ and stratified cross validation to ensure balanced classes in every fold. In Figure \ref{fig:confusion_matrix}, we show the confusion matrix for the 7-crop (all crop types) experiment.\footnote{Code is provided at \url{https://github.com/nasaharvest/cv4a-kenya-knn}.}

Table \ref{results} shows that accuracy was highest for cassava in all experiments (up to 70\% accuracy, comparable with previously published work on smallholder farms, e.g., \citet{Rustowicz2019,Jin2019,Aguilar2018,Lambert2018}). This is consistent with Figure \ref{fig:bandsep}, which shows the NDVI time series for cassava is more distinct from maize and common bean, which have considerable overlap. In general, accuracy for all classes decreases as more classes are added to the dataset. Of the single-crop classes, common bean has the lowest accuracy---the confusion matrix in Figure \ref{fig:confusion_matrix} shows that common bean examples are more similar to examples in other classes than within-class, meaning it is dispersed in the feature space. Intercropped maize/cassava is most similar to single-crop maize and cassava, which is expected since these classes have the highest in-class similarity (evidenced by higher per-class accuracy). 

The results from these experiments show that more information---not more training examples---is needed to increase the separability between crop type classes. While Sentinel-2 acquires images over the dataset locations approximately every 5 days, most of these were not included in the dataset either because they were cloudy or assumed by the dataset authors to be outside of the growing season. The dataset observations cover only half of the main growing season (Figure \ref{fig:ndvi_timeseries}) and contains only one observation during the peak month of June. Increasing the temporal resolution of the NDVI time series and including observations from the full growing season could significantly improve the classification performance. Additional observations could come from optical daily-acquired PlanetScope or 16-day Landsat-8 images as well as SAR Sentinel-1 images, which have been shown to correlate with crop phenology\citep{Mcnairn2016}. More observations in the beginning and peak of the growing season may also reveal more separability in the spectral bands that was not captured in this dataset. Spatial or textural inputs could also improve performance, particularly where high-resolution images might indicate the presence of intercropping. Additional information such survey dates, planting dates, rainfall, temperature, and intercropping percentages could also improve classification, e.g., by enabling fields with failed crops to be removed from the training dataset. However, location information for the Sentinel-2 observations is not provided for the dataset due to privacy concerns and thus precludes the use of any additional data.

\section{Conclusions}
 To promote research in the area of crop type mapping for smallholder agriculture, Radiant Earth Foundation and Plant Village released a new dataset of ground-truth field boundaries and crop types in western Kenya, co-registered with multispectral images from the Sentinel-2 satellites. We provided an interpretation of this dataset in the context of the 2019 growing season, which was made atypical by drought, which should be considered when using this dataset as a benchmark for crop type classification methods generally. In addition, we demonstrated crop type classification performance as high as 64\% for maize and 70\% for cassava using $k$ Nearest Neighbors---a fast, interpretable, scalable method that can be used as a baseline for future methods evaluated on this dataset. This method relies only on the NDVI time series, thus additional data sources (e.g., Landsat-8 and PlanetScope optical or Sentinel-1 SAR images) can readily be integrated to improve separability of crop type classes. However, additional Earth observation data cannot be obtained since location information is not provided for this dataset. While omission of location information is one way to preserve farmer anonymity, it greatly limits the impact of the dataset for advancing research in crop type classification. 

\theendnotes

\bibliography{references}
\bibliographystyle{iclr2020_conference}


\end{document}